\documentclass{article}
\PassOptionsToPackage{hyphens}{url}
\usepackage{iclr2026_conference}
\usepackage{times}
\iclrfinalcopy

\usepackage[colorlinks,linkcolor=blue!60!black,citecolor=blue!60!black,urlcolor=blue!60!black]{hyperref}
\usepackage{url}
\usepackage{graphicx}
\usepackage{caption}

\usepackage{algorithm}
\usepackage{algorithmic}
\usepackage{amsmath}
\usepackage{amssymb}
\usepackage{amsthm}
\usepackage{booktabs}
\usepackage{multirow}
\usepackage{xcolor}
\usepackage{subcaption}
\usepackage{tcolorbox}
\tcbuselibrary{breakable}
\usepackage{wrapfig}

\theoremstyle{definition}

\theoremstyle{plain}

\title{No Time Like the Present: Agentic Test-Time Training for LLM Agents}

\author{%
  Yanbo Wang \quad
  Jinhua Hao\thanks{Corresponding author.} \quad
  Yuze Shi \quad
  Kun Yuan \quad
  Ming Sun \\[0.4em]
  Kuaishou Technology
}

\begin{document}

\maketitle
\lhead{Preprint}

\begin{abstract}
LLM agents often degrade over long episodes: as trajectories grow, they revisit explored states, repeat failed actions, and lose strategies that previously worked. Test-time training (TTT) offers a way to adapt model weights to the evolving task state, but existing LLM TTT methods largely adapt once to a fixed input. We study continuous TTT in multi-turn agent episodes, where each update changes the policy that generates later training text. This creates a self-training loop that helps when new trajectory information appears, but can amplify drift when the agent gets stuck and repeatedly trains on similar text. We find that update-text repetition distinguishes these regimes and introduce Agentic Test-Time Training (aTTT), a token-level reweighting method that downweights the loss on tokens appearing in repeated $n$-grams from prior updates while leaving novel tokens fully weighted. To run such updates inside live episodes, we build a concurrent serving system using vLLM's runtime LoRA API, limiting overhead to 1.9$\times$ the no-TTT cost. aTTT improves success by up to 5.0 points on ALFWorld and 4.9 points on SWE-bench Lite. The gains concentrate where models already have task competence but drift over long trajectories, suggesting that aTTT mainly preserves existing competence rather than teaching new abilities.
\end{abstract}

%===============================================================
\section{Introduction}
\label{sec:intro}
%===============================================================

LLM agents that interact with environments over many turns face a failure mode that worsens with trajectory length. As episodes grow, agents often revisit explored states, repeat failed actions, or explore so slowly that they stop making progress. These failures do not always indicate that the task is beyond the model's competence. Many occur on instances that the model has a nontrivial chance of solving, but where a particular long rollout drifts away from useful evidence and previously promising strategies. The problem is therefore not simply whether the model can solve the task in principle, but whether it can preserve and act on that potential as the trajectory unfolds.

Several strategies address parts of this problem, but none directly targets within-episode drift. Reflexion \citep{shinn2023reflexion} and related methods revise prompts across trials, leaving the agent unchanged within a single episode. Longer context windows expose the full trajectory to the model, but the trajectory itself can become difficult to use as it grows, and models increasingly fail to retrieve or act on earlier evidence \citep{liu2024lost}. Decoding-time repetition penalties can suppress repeated surface tokens, but they do not adapt the policy that keeps revisiting the same states or taking the same actions. We therefore seek a mechanism that adapts the agent during the episode, especially on trajectories where the model retains task-solving potential but begins to lose it through repetition and slow exploration.

\begin{figure}[t]
\vspace{-16pt}
\centering
\includegraphics[width=0.91\textwidth]{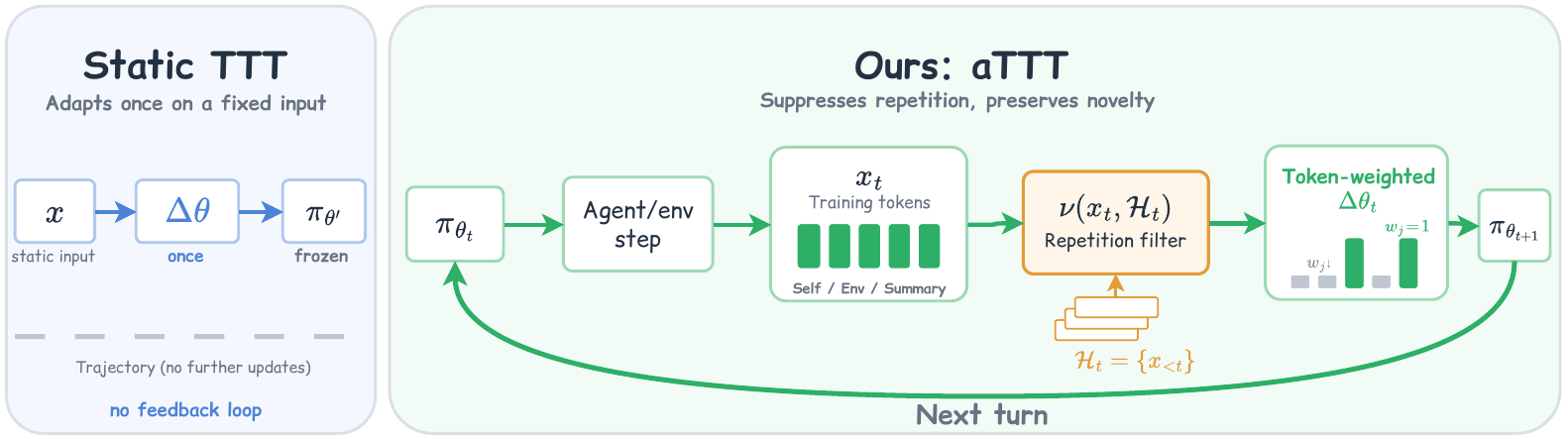}
\caption{Static TTT and aTTT differ in when adaptation occurs. Static TTT adapts once to a fixed input and then freezes the update, leaving the later trajectory outside the training loop. aTTT instead updates during the episode: each agent-environment step provides training text, repeated content is filtered through token weighting, and the updated adapter is used in later turns.}
\label{fig:teaser}
\vspace{-1em}
\end{figure}

Agent trajectories provide a natural signal for test-time training. Each episode produces a task-specific stream of actions, observations, and feedback, which records what the agent has tried, where it has made progress, and where it begins to drift. Static TTT methods such as qTTT \citep{bansal2025let}, In-Place TTT \citep{feng2026place}, and ETT \citep{zahirnia2025ett} show that adapting to test-time text can improve language-model behavior, but they adapt to a fixed input and then stop. Agent episodes are different. The input grows over time, and the model's own outputs can become part of the future training stream. This growing, self-generated training stream motivates our focus on continuous test-time training in multi-turn agent episodes, where adaptation and data generation are coupled throughout the rollout.

Continuous adaptation creates a risk that does not appear in static TTT. At each update, the current policy helps produce the text used for training. The resulting update then changes the policy that produces later text. This feedback can help when the agent keeps discovering new information, because later updates carry fresh evidence from the trajectory. It can also hurt when the agent gets stuck. In that case, later updates often replay actions, observations, or reasoning patterns that earlier updates have already reinforced. The update stream then strengthens the behavior that prevents progress. We call this failure mode the \emph{repeated self-training loop}, as illustrated in Figure~\ref{fig:teaser}.

This failure becomes visible through update-text repetition. As shown in Figure~\ref{fig:novelty}, successful trajectories continue to introduce new content, while failed trajectories show an early and sustained rise in repetition. We use this observation to introduce Agentic Test-Time Training(aTTT), a token-level reweighting method for multi-turn TTT. Instead of dropping an entire repeated update, aTTT downweights the loss on tokens that appear in repeated $n$-grams from prior updates while keeping novel tokens fully weighted. This lets the agent preserve new trajectory information without repeatedly training on the same stuck patterns.

Across ALFWorld and SWE-bench Lite, aTTT improves over static pre-rollout adaptation and unfiltered in-episode training. As reported in Table~\ref{tab:main}, aTTT improves ALFWorld success rates by up to 5.0 points. On SWE-bench Lite, the gain reaches 4.9 points. The gains concentrate in settings where the model has a meaningful chance of solving the task but loses progress over long trajectories. This supports our view that aTTT mainly stabilizes existing task-solving potential rather than teaching new abilities from scratch. To make this practical, we build a concurrent serving system on vLLM's runtime LoRA API \citep{kwon2023efficient}, which serves 16 episodes simultaneously, runs 6.6$\times$ faster than a sequential implementation, and keeps wall-clock cost to 1.9$\times$ the no-TTT baseline.

Our contributions can be listed as follows:
\begin{enumerate}\itemsep2pt
\item We formulate continuous test-time training for multi-turn LLM agents, where the model is updated throughout an episode rather than once on a fixed input. This setting introduces an endogenous feedback loop in which each update can influence the text used for later updates.

\item We identify update-text repetition as a simple signal of when this feedback loop becomes harmful, and propose aTTT. aTTT downweights the training loss on tokens appearing in repeated $n$-grams from prior updates while keeping novel tokens fully weighted.

\item We show that aTTT improves agent performance on ALFWorld and SWE-bench Lite, with gains concentrated in settings where models have task-solving potential but drift over long trajectories. This suggests that aTTT mainly stabilizes existing competence rather than teaching new abilities from scratch.
\end{enumerate}

%===============================================================
\section{Related Work}
\label{sec:related}
%===============================================================

\paragraph{Test-time training and adaptation.}
Test-time training adapts model parameters at inference time using signals available from the test input \citep{sun2020test}. Existing methods instantiate this idea through entropy minimization \citep[TENT,][]{wang2020tent}, input augmentation \citep[MEMO,][]{zhang2022memo}, masked autoencoding \citep{gandelsman2022test}, and analyses of when self-supervised test-time objectives help \citep{liu2021ttt++}. A separate line of work studies the risk of continual adaptation, where repeated updates can accumulate errors or collapse. Methods such as EATA \citep{niu2022efficient}, CoTTA \citep{wang2022continual}, and SAR \citep{niu2023towards} address this risk through sample selection, weight restoration, or reliability filtering. These methods generally adapt on externally supplied test examples. In contrast, multi-turn agents create an endogenous setting: the adapted policy helps generate the future text that will be used for later updates.

\paragraph{TTT for language models.}
Recent work extends TTT to language models. Hardt and Sun \citep{hardt2024test} combine TTT with nearest-neighbor retrieval. qTTT \citep{bansal2025let} improves long-context retrieval by updating query projections, ETT \citep{zahirnia2025ett} expands long-context capacity through chunked adaptation, and In-Place TTT \citep{feng2026place} performs efficient chunk-wise next-token adaptation. TTT has also been studied as a mechanism for few-shot learning \citep{akyurek2024surprising}. These methods adapt to a fixed text input or a fixed context before using the adapted model. Our setting is different because the input is a growing agent trajectory, and later training text can depend on earlier adapted outputs.

\paragraph{Test-time adaptation of LLM agents.}
Closest to our setting, \citet{chen2026test} and \citet{acikgoz2025self} also adapt LLM agents at test time. Their goals are complementary to ours. \citet{chen2026test} adapt agents to unfamiliar environment formats and transition dynamics, while \citet{acikgoz2025self} generate synthetic training examples from uncertain cases to improve missing capabilities. We instead focus on within-episode drift in tasks where the model has nontrivial task-solving potential but may lose progress over a long rollout. The challenge is not primarily environment mismatch or capability acquisition, but the feedback loop created when repeated agent behavior becomes repeated training data. Reflexion \citep{shinn2023reflexion} and related prompting methods revise behavior across trials; our method updates weights within a single trial.

\paragraph{Long-context degradation and self-generated collapse.}
Our work also relates to long-context degradation and collapse under self-generated data. Lost-in-the-middle \citep{liu2024lost} shows that models can fail to use information in long contexts. In agent episodes, the relevant evidence may remain in the trajectory, but the policy may stop using it reliably as the rollout grows. Separately, training on self-generated data can cause distributional collapse \citep{shumailov2024ai}, self-distillation can amplify errors \citep{mobahi2020self}, and policy entropy can collapse during reinforcement learning \citep{cui2025entropy}. These phenomena usually unfold over many training iterations. We study a compressed version at test time, where repeated actions, observations, or reasoning patterns can return as repeated update text within a single episode.

%===============================================================
\section{Agentic Test-Time Training}
\label{sec:method}
%===============================================================

We study test-time training inside a single agent episode. Unlike static TTT, an agent episode produces a growing trajectory whose later content depends on the current policy. Once the policy is updated, it can influence the actions, observations, and text that appear in later updates. The training stream is therefore endogenous: it is partly shaped by the model being trained. This creates the central challenge for multi-turn TTT. When the agent is making progress, new update text can carry useful trajectory information. When the agent gets stuck, the update stream can replay the same actions, observations, or reasoning patterns, causing later updates to reinforce behavior that has already failed. We introduce \emph{Agentic Test-Time Training} (aTTT), which updates an episode-specific adapter during the rollout while reducing the loss contribution of tokens that appear in repeated $n$-grams from earlier updates.

\begin{figure}[t]
\centering
\includegraphics[width=0.82\textwidth]{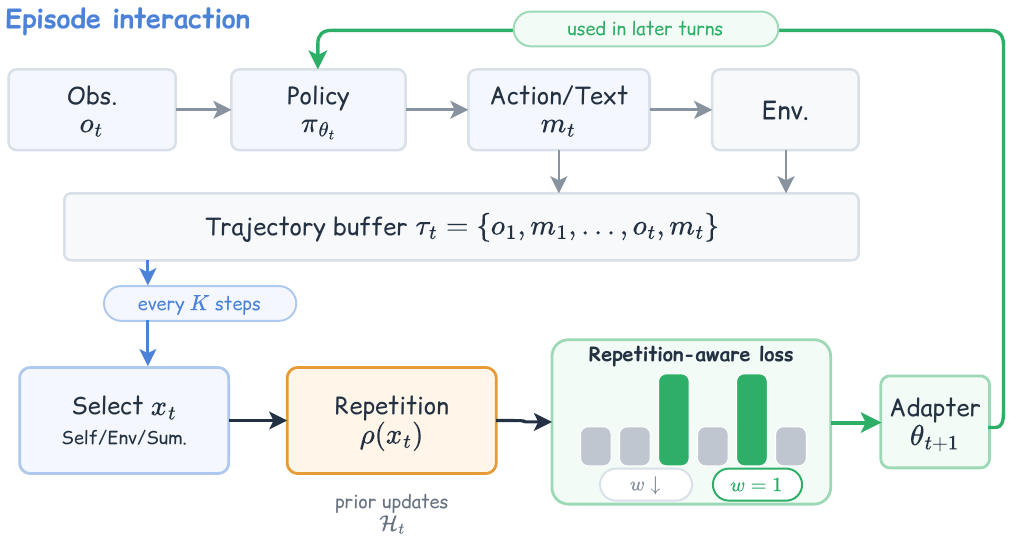}
\caption{Overview of aTTT. Every $K$ steps, the trajectory buffer provides update text from Self, Env, or Summary signals. Previous update texts are used to measure repetition and compute token-level $n$-gram exposure. The repetition-aware loss downweights repeated spans, updates an episode-specific adapter, and the adapted policy is used in later turns.}
\label{fig:method}
\end{figure}

Figure~\ref{fig:method} gives an overview of the online update loop. The rest of this section first formalizes continuous TTT for agent trajectories and the update signals we consider, then defines update-text repetition as a diagnostic for harmful self-reinforcement, and finally presents the token-level reweighting objective used by aTTT.

\subsection{Online TTT for Agent Episodes}

Consider an agent with policy $\pi_\theta$ interacting with an environment over $T$ steps. At each environment step $t$, the agent receives an observation $o_t$ and produces a message $m_t = (r_t, u_t)$, where $r_t$ is the reasoning trace and $u_t$ is the executable action. The trajectory up to step $t$ is
\begin{equation}
\tau_t = (o_1, m_1, o_2, m_2, \ldots, o_t, m_t).
\label{eq:traj}
\end{equation}

Let $k$ index test-time updates, and let $t_k$ be the environment step at which the $k$-th update is performed. At update $k$, we select a candidate text $x_k$ from the current trajectory $\tau_{t_k}$ and adapt the policy with next-token prediction:
\begin{align}
\theta_{k+1} &= \theta_k - \eta\,\nabla_{\theta_k} \mathcal{L}_{\text{NTP}}(x_k,\theta_k), \label{eq:update}\\
\mathcal{L}_{\text{NTP}}(x_k,\theta_k)
&= -\textstyle\sum_i \log \pi_{\theta_k}(x_{k,i} \mid x_{k,<i}). \nonumber
\end{align}

The source of $x_k$ determines what information enters the update stream. We study three training signals:
\begin{align}
x_k^{\text{Self}} &= m_{t_k}, \label{eq:self}\\
x_k^{\text{Env}} &= o_{t_k}, \label{eq:obs}\\
x_k^{\text{Summary}} &= g(\tau_{t_k}). \label{eq:digest}
\end{align}
Self trains on the agent's own reasoning and action tokens, so it has the most direct path from model output to future training text. Env trains on environment-generated observations and does not directly replay the model's output. Summary trains on a compressed progress note produced by a separate LLM call $g(\tau_{t_k})$, which aggregates information across the trajectory at the cost of one additional inference step.

For Self and Env, we use only the most recent step rather than the full prefix. This keeps the update focused on the current state and avoids repeatedly training on the entire early trajectory. The adapter persists within the episode, so earlier updates continue to affect later turns.

\subsection{Update-Text Repetition}

Online TTT introduces a feedback loop that is absent from single-input adaptation. In static TTT, the training text is fixed before the update. In agent episodes, the current policy helps produce the trajectory, the trajectory supplies update text, and the updated policy then produces later trajectory content. Each update can therefore influence the text used for future updates.

\begin{wrapfigure}{r}{0.42\textwidth}
\centering
\includegraphics[width=\linewidth]{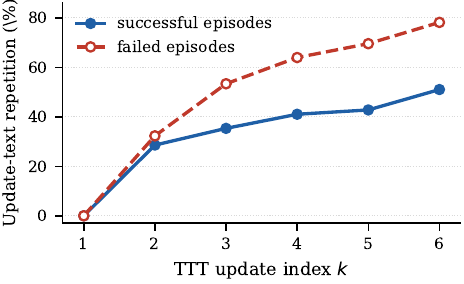}
\caption{Update-text repetition as a diagnostic signal. On Qwen3.5-9B with the Self signal, failed episodes show an early and sustained rise in repetition across TTT updates.}
\label{fig:novelty}
\vspace{-8pt}
\end{wrapfigure}

This loop has two regimes. When the agent is making progress, later update texts contain new observations, actions, and state information. Updating on this text can help the model retain useful trajectory evidence as the context grows. When the agent gets stuck, later update texts often repeat actions, observations, or reasoning patterns that earlier updates have already reinforced. Updating again on the same patterns can strengthen the behavior that prevents progress.

We use \emph{update-text repetition} as an operational signal of the stuck regime. Let $x_k$ be the candidate text for the $k$-th update, and let $\mathcal{H}_k = \{x_1,\ldots,x_{k-1}\}$ be the set of previous update texts in the same episode. We define repetition as the largest word-level Jaccard overlap between the current update text and any previous update text:
\begin{equation}
\rho(x_k) = \max_{x' \in \mathcal{H}_k} J\!\bigl(W(x_k),\, W(x')\bigr),
\label{eq:repetition}
\end{equation}
where $W(x)$ is the set of whitespace-tokenized words in $x$ and $J$ is Jaccard similarity. For the first update in an episode, $\mathcal{H}_k$ is empty and we set $\rho(x_k)=0$. High repetition means that the candidate update largely replays text used by an earlier update.

We also define the corresponding novelty score
\begin{equation}
\nu(x_k) = 1 - \rho(x_k).
\label{eq:novelty}
\end{equation}
We use this score in Section~\ref{sec:experiments} to build a sequence-level baseline that skips an update when $\nu(x_k)$ falls below a threshold.

As shown in Figure~\ref{fig:novelty}, update-text repetition separates successful and failed trajectories early in the episode. Successful episodes continue to produce comparatively novel update text, while failed episodes show an early and sustained rise in repetition. This supports using repetition as a diagnostic signal that a candidate update may contain little new information and may instead reinforce an existing stuck pattern. We use repetition in this operational sense, without assuming that repetition alone causes failure.

\subsection{Repetition-Aware Token Reweighting}
\label{sec:token_reweighting}

A repeated update is rarely entirely useless. It may contain a repeated reasoning template together with a new observation, object name, location, or action. A sequence-level filter that drops the whole update can remove these useful tokens along with the repeated text. aTTT therefore keeps each update but changes how much each token contributes to the loss.

Let $\mathcal{H}^{\text{tok}}_k$ be the concatenation of all token sequences used in previous updates within the same episode. Tokenize the current update text as $x_k = [x_{k,1}, \ldots, x_{k,L}]$. For each token position $j$, we compute its maximum $n$-gram exposure in the previous update history:
\begin{equation}
f_k(j) =
\max_{g \ni j}
\text{count}\bigl(g, \mathcal{H}^{\text{tok}}_k\bigr),
\label{eq:freq}
\end{equation}
where the maximum is over all $n$-grams in $x_k$ that contain position $j$. We then assign the token-level loss weight
\begin{equation}
w_{k,j} =
\max\!\left(w_{\min},\, \frac{1}{1 + f_k(j)}\right).
\label{eq:weight}
\end{equation}

Tokens that do not appear in repeated $n$-grams keep weight $1$. Tokens inside frequently repeated $n$-grams receive smaller weights, down to the floor $w_{\min}$. The repetition-aware loss is
\begin{equation}
\mathcal{L}_{\text{aTTT}}(x_k,\theta_k)
=
\frac{1}{L}
\sum_{j=1}^{L}
w_{k,j}\,
\text{CE}(x_{k,j}, \hat{x}_{k,j}; \theta_k).
\label{eq:attt_loss}
\end{equation}
Because repeated tokens have smaller loss weights, their gradient contribution is reduced. Novel tokens remain fully weighted and can still update the adapter. Algorithm~\ref{alg:gate} gives the full procedure.

\begin{algorithm}[t]
\caption{aTTT update}
\label{alg:gate}
\textbf{Input}: update text $x_k$, token history $\mathcal{H}^{\text{tok}}_k$, $n$-gram size $n$, floor $w_{\min}$
\begin{algorithmic}[1]
\STATE Tokenize $x_k$ into $[x_{k,1}, \ldots, x_{k,L}]$
\FOR{$j = 1$ to $L$}
    \STATE $f_k(j) \leftarrow \max_{g \ni j} \text{count}(g, \mathcal{H}^{\text{tok}}_k)$
    \STATE $w_{k,j} \leftarrow \max(w_{\min},\, 1/(1+f_k(j)))$
\ENDFOR
\STATE $\theta_{k+1} \leftarrow \theta_k - \eta \nabla_{\theta_k}
\frac{1}{L}\sum_{j=1}^{L} w_{k,j} \cdot \text{CE}(x_{k,j}, \hat{x}_{k,j}; \theta_k)$
\STATE $\mathcal{H}^{\text{tok}}_{k+1} \leftarrow \mathcal{H}^{\text{tok}}_k \,\|\, [x_{k,1}, \ldots, x_{k,L}]$
\end{algorithmic}
\end{algorithm}

%===============================================================
\section{Experiments}
\label{sec:experiments}
%===============================================================

We evaluate aTTT as an online adaptation method for multi-turn agents. The experiments ask whether in-episode updates improve over no adaptation and static pre-rollout TTT, whether repetition-aware token reweighting improves over simpler filtering baselines, and when the gains appear across models, task types, and trajectory lengths.

\subsection{Experimental Setup}

\paragraph{Datasets.}
We evaluate on ALFWorld \citep{shridhar2020alfworld} and SWE-bench Lite \citep{jimenez2024swe}. ALFWorld is a household instruction-following benchmark with immediate environment feedback and a 50-step budget. SWE-bench Lite is a software-engineering benchmark with longer trajectories and deferred verification. Appendix~\ref{app:bench} reports the details of experiment settings.

\paragraph{Models.}
We evaluate Qwen3.5-4B, Qwen3.5-9B, and Qwen3.5-27B \citep{qwen35blog}, and Gemma-3-12B \citep{gemmateam2025gemma3technicalreport}. These models span different baseline success rates and failure profiles, which lets us test whether aTTT helps uniformly or only under particular competence and horizon regimes. Training details, seeds, and hyperparameters are in Appendix~\ref{app:signals}.

\paragraph{Compared methods.}
We compare several adaptation and control settings. ReAct is the no-TTT baseline. qTTT \citep{bansal2025let} is static pre-rollout adaptation. No filter denotes online TTT with the same in-episode update schedule as aTTT but with the vanilla next-token prediction loss. Sequence filter uses the update-text novelty score from Section~\ref{sec:method} to skip highly repeated updates entirely. aTTT is our online TTT method, which keeps each selected update but applies repetition-aware token-level loss reweighting. For online methods, the update text comes from one of three sources: Self, Env, or Summary. We also evaluate non-parametric controls, including in-context delivery of the same Summary text and decoding-time repetition penalties, in Section~\ref{sec:controls}.

\paragraph{Metrics and protocol.}
For both benchmarks, we report success rate, averaged over three seeds with standard error. Unless otherwise noted, aTTT uses a LoRA adapter that is reset between episodes and persists within an episode. The default update cadence is $K{=}5$ agent steps.

\paragraph{System and throughput.}
Online TTT requires many in-episode updates. We decouple inference from training with per-episode LoRA adapters served through vLLM's runtime LoRA API \citep{kwon2023efficient}. The system runs 16 episodes concurrently and keeps wall-clock cost to 1.9$\times$ the no-TTT run Architecture and throughput details are in Appendix~\ref{app:system}.

\subsection{Main Results}

\begin{table}[t]
\centering
\small
\setlength{\tabcolsep}{4.5pt}
\caption{ALFWorld success rates (\%), mean$\pm$SE over three seeds. Bold indicates the best filter within each signal-model block.}
\label{tab:main}
\begin{tabular}{ll ccc}
\toprule
\textbf{Signal} & \textbf{Filter} & \textbf{Qwen3.5-4B} & \textbf{Gemma-3-12B} & \textbf{Qwen3.5-9B} \\
\midrule
\multicolumn{2}{l}{ReAct} & $1.9_{\pm0.5}$ & $22.4_{\pm0.6}$ & $50.7_{\pm0.8}$ \\
\multicolumn{2}{l}{qTTT} & $2.6_{\pm0.6}$ & $22.6_{\pm1.7}$ & $52.4_{\pm1.4}$ \\
\midrule
\multirow{3}{*}{Self} & No filter & $2.1_{\pm0.4}$ & $22.1_{\pm1.1}$ & $53.1_{\pm1.9}$ \\
 & Sequence filter & $3.1_{\pm1.0}$ & $23.8_{\pm0.9}$ & $53.8_{\pm1.7}$\\
 & aTTT & $\mathbf{3.6}_{\pm1.1}$ & $\mathbf{24.3}_{\pm1.1}$ & $\mathbf{55.7}_{\pm1.4}$\\
\cmidrule{1-5}
\multirow{3}{*}{Env} & No filter & $4.3_{\pm0.4}$ & $21.7_{\pm1.3}$ & $51.0_{\pm1.0}$ \\
 & Sequence filter & $\mathbf{5.5}_{\pm0.5}$ & $22.9_{\pm1.1}$ & $52.1_{\pm1.1}$ \\
 & aTTT & $4.5_{\pm1.0}$ & $\mathbf{25.0}_{\pm1.1}$ & $\mathbf{55.5}_{\pm1.3}$ \\
\cmidrule{1-5}
\multirow{3}{*}{Summary} & No filter & $4.5_{\pm0.5}$ & $25.5_{\pm1.3}$ & $52.1_{\pm0.8}$ \\
 & Sequence filter & $4.3_{\pm0.4}$ & $26.2_{\pm1.4}$ & $54.0_{\pm1.7}$ \\
 & aTTT & $\mathbf{4.8}_{\pm0.2}$ & $\mathbf{26.7}_{\pm0.6}$ & $\mathbf{54.3}_{\pm0.7}$ \\
\bottomrule
\end{tabular}
\end{table}

\paragraph{Online adaptation needs repetition control.}
Table~\ref{tab:main} shows that adapting only once before the rollout is not sufficient for agent episodes. qTTT stays close to ReAct across models, suggesting that pre-rollout adaptation cannot exploit the trajectory information that appears during interaction. Online TTT has access to this information, but the unfiltered variant is inconsistent. It improves in some settings, while stagnating or degrading in others. This supports our central claim that in-episode updates are useful only when the repeated self-training loop is controlled. Repetition-aware methods generally improve over unfiltered online TTT, showing that update-text repetition is a useful signal for identifying low-novelty updates that may reinforce stuck behavior.

\paragraph{aTTT preserves novelty without external supervision.}

\begin{wrapfigure}{r}{0.42\textwidth}
\vspace{-12pt}
\centering
\includegraphics[width=\linewidth]{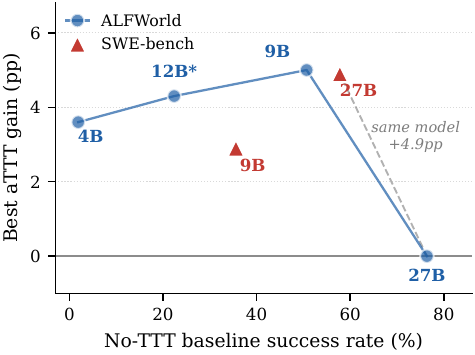}
\caption{aTTT is most effective when baseline task-solving potential and trajectory length coexist. 12B$^*$ denotes Gemma-3-12B.}
\label{fig:sweet}
\vspace{-1em}
\end{wrapfigure}

aTTT is usually stronger than sequence filtering because it avoids an all-or-nothing decision. A sequence filter can remove repeated updates, but it also discards new observations, object names, locations, or action outcomes that appear inside partially repeated text. aTTT keeps each update and only reduces the loss on repeated token spans, leaving novel tokens fully weighted. The gains should also be interpreted in light of the supervision available at test time. aTTT does not use external knowledge, solved demonstrations, or additional task-specific training data. All update text comes from the live episode itself. Therefore, even gains of a few points are meaningful, because they come from stabilizing the agent during inference rather than injecting new task knowledge. The best update source remains model-dependent, with Env strongest for Qwen3.5-4B, Self and Env both strong for Qwen3.5-9B, and Summary strongest for Gemma-3-12B. This shows a limitation of the method. aTTT makes a chosen update stream safer, but does not by itself solve source selection.

\subsection{Cross-Benchmark Evaluation}

\begin{wraptable}{r}{0.48\textwidth}
\vspace{-10pt}
\centering
\small
\setlength{\tabcolsep}{3pt}
\caption{SWE-bench Lite resolve rates (\%), mean$\pm$SE over three seeds.}
\label{tab:swe}
\begin{tabular}{@{}l cc@{}}
\toprule
 & \textbf{Qwen3.5-9B} & \textbf{Qwen3.5-27B} \\
\midrule
No TTT & $35.6_{\pm0.4}$ & $57.8_{\pm0.6}$ \\
aTTT & $\mathbf{38.4}_{\pm0.6}$ {\footnotesize\textcolor{gray}{($+2.9$)}} 
     & $\mathbf{62.7}_{\pm1.0}$ {\footnotesize\textcolor{gray}{($+4.9$)}} \\
\bottomrule
\end{tabular}
\vspace{-1em}
\end{wraptable}

We next evaluate whether the same online adaptation rule remains useful outside ALFWorld. SWE-bench Lite differs in domain, feedback structure, and horizon: the agent must edit software repositories, verification is deferred, and trajectories are substantially longer. We use the Summary signal for aTTT because it provides a compact trajectory-level update source for long software-engineering episodes.

As shown in Table~\ref{tab:swe}, aTTT improves resolve rates for both models, with a 2.9-point gain on Qwen3.5-9B and a 4.9-point gain on Qwen3.5-27B. These results are consistent with the ALFWorld pattern, but they should not be read as universal gains from adaptation. aTTT is most effective when the model has nontrivial task-solving potential and the trajectory is long enough for repeated updates to become low-novelty. Qwen3.5-27B gains little from adaptation on shorter ALFWorld episodes, where it is already relatively stable, but benefits on SWE-bench Lite. For this model, mean update-text repetition on SWE-bench Lite is about six times higher than on ALFWorld.

Figure~\ref{fig:sweet} summarizes this regime across model scales and horizons. Gains are small when the model is too weak to act reliably, and also small when the model is already stable on short trajectories. The largest gains appear when baseline task-solving potential and long-horizon drift coexist.

\subsection{When Does aTTT Help?}

\begin{wrapfigure}{r}{0.46\textwidth}
\vspace{-12pt}
\centering
\includegraphics[width=0.44\textwidth]{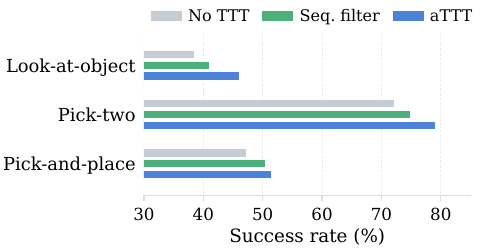}
\caption{ALFWorld success by task type. Gains concentrate on pick-two tasks, where the agent must sustain progress across multiple object searches.}
\label{fig:tasktype}
\vspace{-10pt}
\end{wrapfigure}

\begin{figure}[t]
\vspace{-14pt}
\centering
\includegraphics[width=0.92\textwidth]{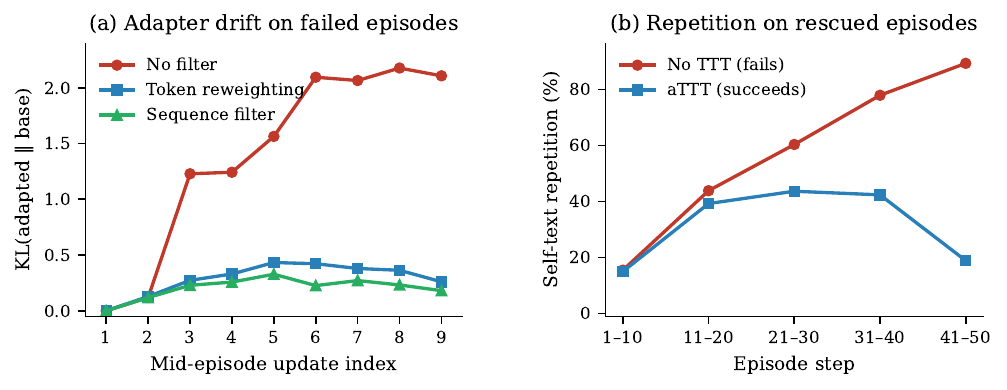}
\caption{Repetition and adapter drift on diagnostic subsets. Left: on failed episodes, unfiltered online TTT moves farther from the base model than aTTT. Right: in rescued pairs, no-TTT trajectories become increasingly repetitive, while aTTT trajectories remain less repetitive before completion.}
\label{fig:drift}
\end{figure}

\paragraph{Repetition and adapter drift.}
We first examine rescued pairs, defined as ALFWorld games that fail without TTT but succeed with aTTT under the same seed. Figure~\ref{fig:drift} shows that aTTT changes the adapter without inducing excessive drift. In the left panel, unfiltered online TTT moves far from the base model on failed episodes, with KL divergence rising above 2.0 in later updates. Sequence filtering is the most conservative because it skips repeated updates entirely. aTTT lies between these two extremes: it makes a non-trivial adapter update, but keeps the KL divergence below 0.5, suggesting that token-level reweighting can adapt the policy while avoiding the large drift caused by repeatedly training on redundant text. The right panel shows the corresponding trajectory-level pattern. In rescued pairs, no-TTT trajectories become increasingly repetitive as the episode progresses, with self-text repetition rising above 80\%. In contrast, aTTT trajectories remain less repetitive through the middle of the episode and decline as the agent completes the task. The two arms have similar repetition in the first 10 steps, and the gap appears after step 21. On average, aTTT-rescued runs finish in 26.3 steps, while the corresponding no-TTT runs reach the 50-step budget. Together, these diagnostics suggest that aTTT helps when the agent begins to repeat low-novelty behavior, by allowing useful adaptation while limiting self-reinforcing drift.

\paragraph{Task type.}
ALFWorld contains 103 pick-and-place games, 24 pick-two games, and 13 look-at-object games. Figure~\ref{fig:tasktype} shows that aTTT improves over no TTT and sequence filtering across all three coarse task types. The gain is also clear on look-at-object tasks, which have the lowest baseline success rate in this split.

\paragraph{Failure signatures and update sources.}
Table~\ref{tab:signature} profiles how the three models fail on ALFWorld and reports the strongest update signal observed in the main sweep. The best signal is model-dependent: Env works best for Qwen3.5-4B, Self is slightly strongest for Qwen3.5-9B with Env close behind, and Summary is strongest for Gemma-3-12B. We do not interpret these patterns as a causal mapping from failure type to signal type. Instead, they show that repetition control alone does not determine performance: aTTT makes a chosen update stream safer, but the usefulness of that stream still depends on the model and trajectory. Selecting or combining update signals online therefore remains a separate open problem. Appendix~\ref{app:case} provides a representative rescued trajectory.

\begin{table}[H]
\centering\small
\setlength{\tabcolsep}{3pt}
\caption{Failure profiles and strongest observed update signal on ALFWorld. The best signal is selected from the main sweep and varies across models.}
\label{tab:signature}
\begin{tabular}{@{}l ccc@{}}
\toprule
 & \textbf{Qwen3.5-4B} & \textbf{Qwen3.5-9B} & \textbf{Gemma-3-12B} \\
\midrule
Unparseable actions & 30\% & 2\% & 6\% \\
Repeats prev.\ action & 71\% & 31\% & 7\% \\
Repeats within last 3 & 87\% & 34\% & 45\% \\
\midrule
Dominant failure & invalid actions & mid-ep.\ loops & slow expl. \\
Best signal & Env & Self & Summary \\
Best gain & $+3.6$ & $+5.0$ & $+4.3$ \\
\bottomrule
\end{tabular}
\vspace{-6pt}
\end{table}

\subsection{Ablations and Controls}
\label{sec:controls}

\begin{figure}[H]
\centering
\includegraphics[width=0.68\textwidth]{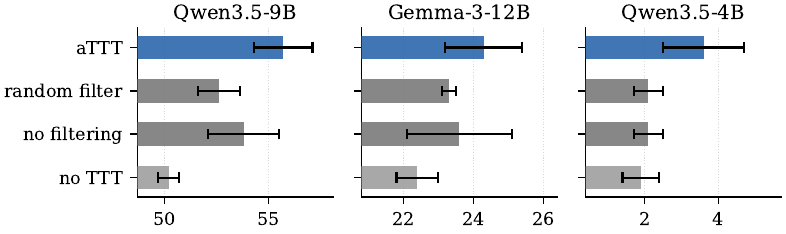}
\caption{Random-drop control. Dropping the same number of updates at random does not match the gain from repetition-based filtering.}
\label{fig:random}
\end{figure}

\paragraph{Is repetition filtering just fewer updates?}
We first test whether the gain comes from filtering repeated updates or simply from training on fewer updates. The random-drop control removes updates at the same empirical rate as the sequence filter, but chooses which updates to drop at random. Figure~\ref{fig:random} shows that random dropping does not match repetition-based filtering. This supports the role of the diagnostic itself: the skipped updates are not arbitrary, but concentrated in parts of the episode where the update stream has become repetitive. For the 4B Self signal, 97\% of late-episode updates are skipped compared to 16\% of early updates.

\paragraph{How sensitive is aTTT to update cadence?}
We update the adapter once every $K$ agent steps. Table~\ref{tab:cadence} reports Qwen3.5-9B on ALFWorld with the Self signal and aTTT, varying $K$ from 1 to 10. Small $K$ gives low novelty and small gains, while $K{=}5$ and $K{=}10$ raise novelty above 0.5 and give larger gains. Cadence controls how often update text enters the stream, while aTTT controls how much repeated spans contribute once an update is selected.

\begin{table}[!htbp]
\centering\small
\setlength{\tabcolsep}{5pt}
\caption{Update cadence ablation (Qwen3.5-9B, Self signal, three seeds).}
\label{tab:cadence}
\begin{tabular}{@{}lccc@{}}
\toprule
\textbf{Cadence $K$} & \textbf{SR (\%)} & \textbf{Mean novelty} & \textbf{vs.\ No TTT} \\
\midrule
$K{=}1$ (every step) & $51.0_{\pm2.6}$ & 0.42 & $+0.3$ \\
$K{=}2$ & $51.9_{\pm4.3}$ & 0.42 & $+1.2$ \\
$K{=}5$ (default) & $55.7_{\pm1.4}$ & 0.56 & $+5.0$ \\
$K{=}10$ & $54.5_{\pm2.9}$ & 0.58 & $+3.8$ \\
\bottomrule
\end{tabular}
\vspace{-6pt}
\end{table}

aTTT uses $n{=}3$ throughout. Appendix~\ref{app:hparam} shows that performance is stable for $n$ between 3 and 5.

\paragraph{Why not put the same text in context?}
We test whether the same trajectory signal can be delivered through the prompt rather than through weight updates. We use the Summary signal, which is the most compact update source and therefore the strongest candidate for in-context delivery. On ALFWorld with Qwen3.5-9B, in-context injection scores 49.3\%, below the no-TTT baseline of 50.7\% and below TTT with the same Summary signal at 54.3\%. This does not imply that prompting cannot use summaries in general, but it shows that, under matched update text, prompt delivery does not reproduce the benefit of adapter updates. One likely reason is that injected summaries consume prompt budget and must compete with the growing trajectory, while adapter updates persist without adding prompt tokens. Details are in Appendix~\ref{app:incontext}.

\paragraph{Why not use decoding-time repetition penalties?}
We also test whether a decoding-time repetition penalty can address the same failure mode. On Qwen3.5-9B with the Self signal, a standard repetition penalty of 1.1 averages 49.7\% across three seeds, below both the no-TTT baseline at 50.7\% and aTTT at 55.7\%. The gap suggests that the relevant repetition is not only a surface-token problem. Decoding penalties discourage recently generated tokens within the response, but they do not update the policy that selects actions across the episode. aTTT instead changes the adapter state used in later turns, allowing trajectory information to affect subsequent action probabilities.

\paragraph{What does token reweighting preserve?}
The advantage of aTTT over sequence filtering is most visible when an update is only partly redundant. A low-novelty text can still contain a new observation, object name, location, or action outcome inside a repeated template. Sequence filtering drops the whole update in this case. Token-level reweighting instead keeps those useful tokens available while reducing the loss on repeated spans. This is why aTTT can improve over an all-or-nothing filter, especially for Env updates where repeated observation framing is often mixed with new state information.

%===============================================================
\section{Discussion}
\label{sec:discussion}
%===============================================================

\paragraph{Redundancy is not relevance.}
aTTT controls redundant self-reinforcement once an update source has been chosen, but it does not decide which source is most useful for a given model or trajectory. This limitation appears in the ALFWorld results: the best update source varies across models, and no single source is uniformly strongest. In particular, Qwen3.5-4B benefits most from Env, Qwen3.5-9B from Self with Env close behind, and Gemma-3-12B from Summary. Selecting or combining update signals online remains an open problem.

%===============================================================
\section{Conclusion}
\label{sec:conclusion}
%===============================================================

Continuous TTT inside multi-turn agent episodes creates an endogenous feedback loop, where an adapted policy can shape the text used for later updates. We introduced aTTT, which uses update-text repetition to downweight the loss on tokens appearing in repeated $n$-grams while leaving novel tokens fully weighted. Across ALFWorld and SWE-bench Lite, aTTT improves performance most in settings where models have nontrivial task-solving potential but drift over long trajectories. These results suggest that in-episode weight updates can stabilize useful agent behavior without acting as broad capability training.

\bibliography{references}
\bibliographystyle{iclr2026_conference}

\clearpage
\appendix

\begin{figure}[H]
\centering
\includegraphics[width=0.82\textwidth]{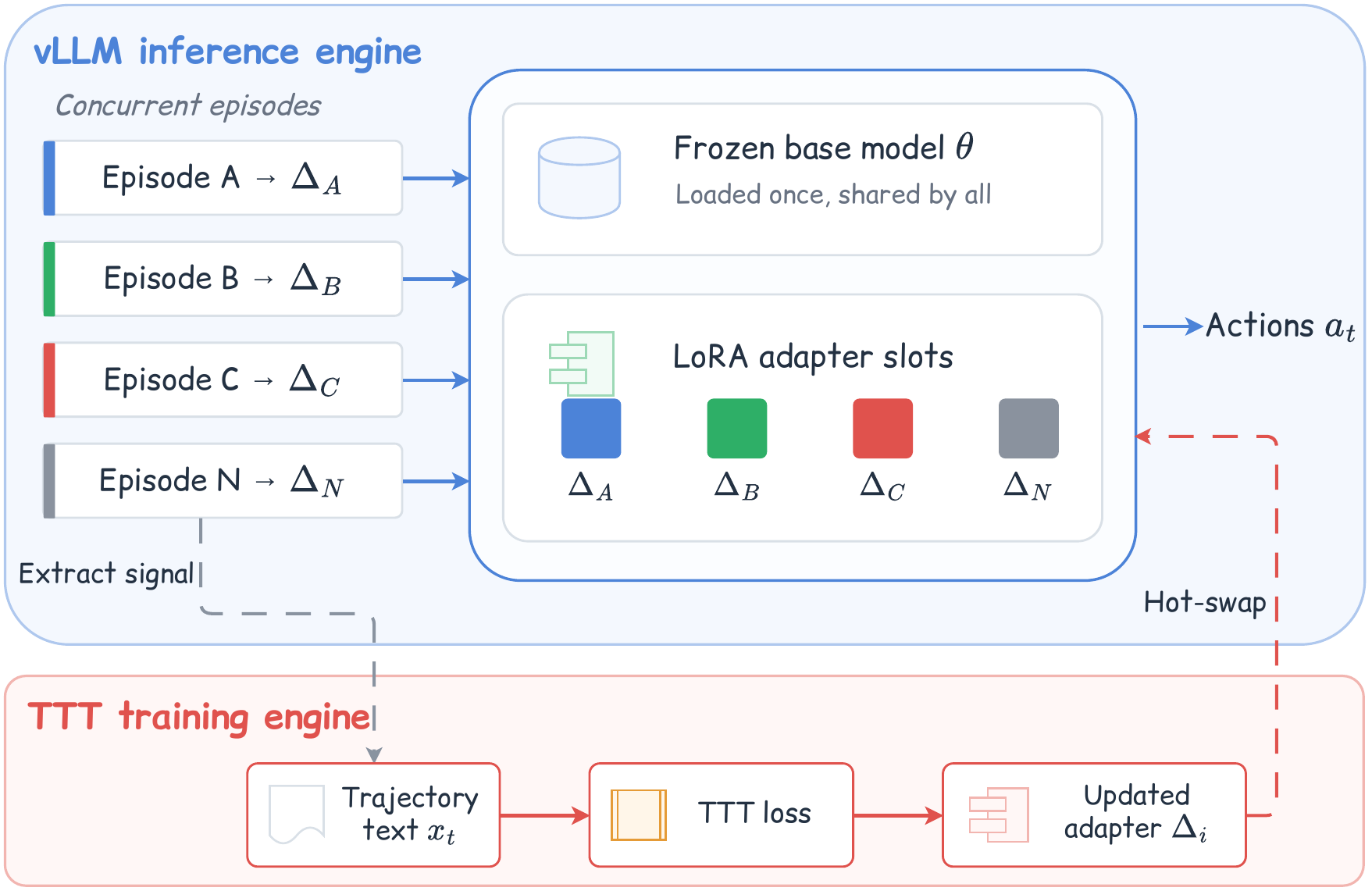}
\caption{Concurrent serving pipeline for online TTT. Each episode maintains a private LoRA adapter, while a separate training engine computes updates and hot-swaps adapters into the vLLM inference engine without blocking other episodes.}
\label{fig:infra}
\end{figure}

\section{System Architecture and Throughput}
\label{app:system}

aTTT requires repeated private updates while many agent episodes remain active. Our implementation decouples training from inference using vLLM's runtime LoRA API \citep{kwon2023efficient}. A frozen base model serves 16 episodes concurrently, each with a private LoRA adapter slot. Dedicated training GPUs compute updates asynchronously and hot-swap the resulting adapter into the serving engine in under 100\,ms. An update to one episode does not block generation for the others. Figure~\ref{fig:infra} sketches the architecture, and Table~\ref{tab:throughput} reports throughput on ALFWorld with Qwen3.5-9B.

\begin{table}[H]
\centering\small
\setlength{\tabcolsep}{5pt}
\caption{Wall-clock cost of 140 ALFWorld episodes. Concurrent aTTT costs 1.9$\times$ the no-TTT run and is 6.6$\times$ faster than a sequential implementation.}
\label{tab:throughput}
\begin{tabular}{@{}lcc@{}}
\toprule
\textbf{Configuration} & \textbf{Wall-clock} & \textbf{vs.\ No TTT} \\
\midrule
No TTT  & 14.6 min & 1.0$\times$ \\
Ours & 28.3 min & 1.9$\times$ \\
Sequential & 186 min & 12.7$\times$ \\
\bottomrule
\end{tabular}
\end{table}

\section{Benchmark and Scaffold Details}
\label{app:bench}

ALFWorld \citep{shridhar2020alfworld} contains 140 text-based household tasks, each with a 50-step budget. We use ReAct-style prompting \citep{yao2022react} on Qwen3.5-4B, Qwen3.5-9B, Qwen3.5-27B, and Gemma-3-12B.

We evaluate SWE-bench Lite \citep{jimenez2024swe} on 150 instances. The agent uses a mini-swe-agent scaffold \citep{yang2024sweagent} with bash tools and a 150-call budget per issue. We run Qwen3.5-9B and Qwen3.5-27B on this benchmark.

\section{Signal Construction and Training Details}
\label{app:signals}

The three training signals expose different parts of the repeated update loop. At each scheduled update, \textbf{Self} uses the model's latest reasoning and action tokens $m_t$. \textbf{Env} uses the latest observation $o_t$ verbatim. \textbf{Summary} prompts the base model to compress the trajectory into a short progress note. Self and Env train only on the most recent step, not the full prefix, so they reflect the current state without replaying the entire trajectory.

Unless noted otherwise, TTT uses a LoRA adapter of rank 8 and $\alpha{=}16$, a learning rate of $5\times10^{-4}$, two gradient steps per update, and a cadence of $K{=}5$ agent steps. The adapter persists within an episode and is reset between episodes. The sequence-level filter uses Jaccard similarity over whitespace-tokenized words with threshold $\tau{=}0.5$, and the token-level reweighting uses $3$-grams with a weight floor $w_{\min}{=}0.05$.

\section{Hyperparameter Sensitivity}
\label{app:hparam}

Table~\ref{tab:hparam} reports a LoRA rank sweep on Qwen3.5-27B across 140 ALFWorld games. Performance is stable across ranks 4 to 16 and degrades modestly at rank 32. We use rank 8 throughout as a conservative default.

\begin{table}[h]
\centering\small
\caption{LoRA rank sensitivity on Qwen3.5-27B. Performance is robust across ranks 4--16.}
\label{tab:hparam}
\begin{tabular}{@{}lc@{}}
\toprule
\textbf{LoRA rank} & \textbf{$\Delta$pp vs baseline} \\
\midrule
4 & $-0.7$ \\
8 (default) & $\pm0$ \\
16 & $-1.4$ \\
32 & $-4.3$ \\
\bottomrule
\end{tabular}
\end{table}

\begin{table}[h]
\centering\small
\caption{$N$-gram window size sensitivity (Qwen3.5-9B, Summary signal). Robust across $n{=}3$--5; we use $n{=}3$ throughout.}
\label{tab:ngram}
\begin{tabular}{@{}lc@{}}
\toprule
\textbf{$n$-gram size} & \textbf{SR (\%)} \\
\midrule
$n{=}2$ & $52.8_{\pm1.8}$ \\
$n{=}3$ (default) & $54.3_{\pm0.7}$ \\
$n{=}4$ & $54.5_{\pm0.6}$ \\
$n{=}5$ & $54.1_{\pm0.8}$ \\
\bottomrule
\end{tabular}
\end{table}

\section{In-Context Baseline Details}
\label{app:incontext}

To test whether weight updates are necessary, we construct an in-context baseline that delivers the same Summary text through the prompt rather than through a LoRA update. At each scheduled update point, the Summary is appended to the system message as a structured plan note. The agent then generates its next action conditioned on this extended context.

On Qwen3.5-9B, in-context injection scores 49.3\%, below the 51.4\% no-TTT baseline. The same Summary text used for TTT weight updates yields 54.3\%. The in-context version fails because the injected plan text competes for attention with the growing trajectory. On episodes longer than 30 steps, the injected note is displaced by more recent observations, and the agent reverts to the same stuck patterns seen in the no-TTT condition. Weight-level internalization avoids this competition by compressing the task-relevant pattern into adapter parameters that persist regardless of context length.

\section{Policy-Level Diagnostics}
\label{app:mechanism}

\begin{figure}[H]
\centering
\includegraphics[width=0.9\textwidth]{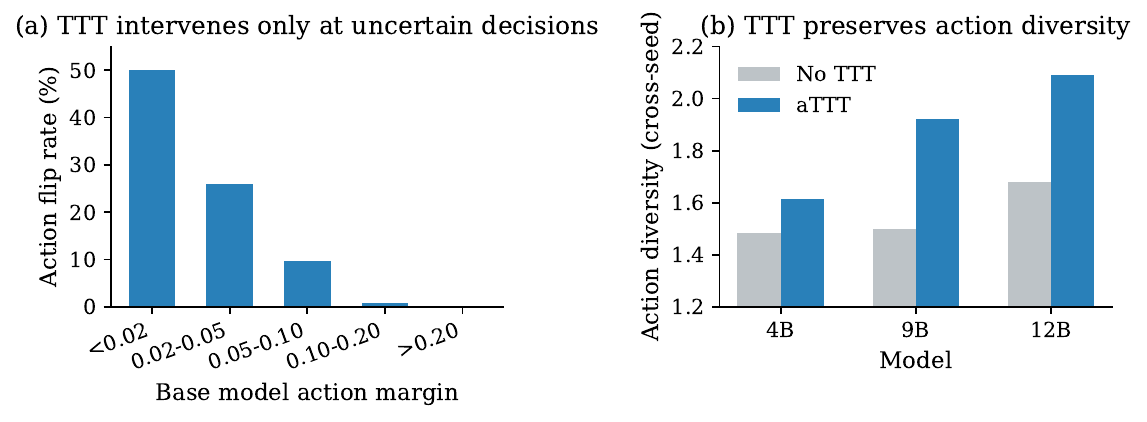}
\caption{Policy-level diagnostics. Action changes concentrate in low-margin decisions, and aTTT increases cross-seed action diversity across models.}
\label{fig:mechanism}
\end{figure}

Figure~\ref{fig:mechanism} provides a supplementary view of how aTTT affects action selection. Action changes concentrate in low-margin decisions, where the base model is less certain, and aTTT increases cross-seed action diversity across models. These diagnostics are consistent with the drift results in the main text: aTTT tends to alter uncertain choices in repeated regimes rather than broadly rewriting high-confidence behavior.

\section{Case Study: Step-by-Step Comparison}
\label{app:case}

\begin{figure}[H]
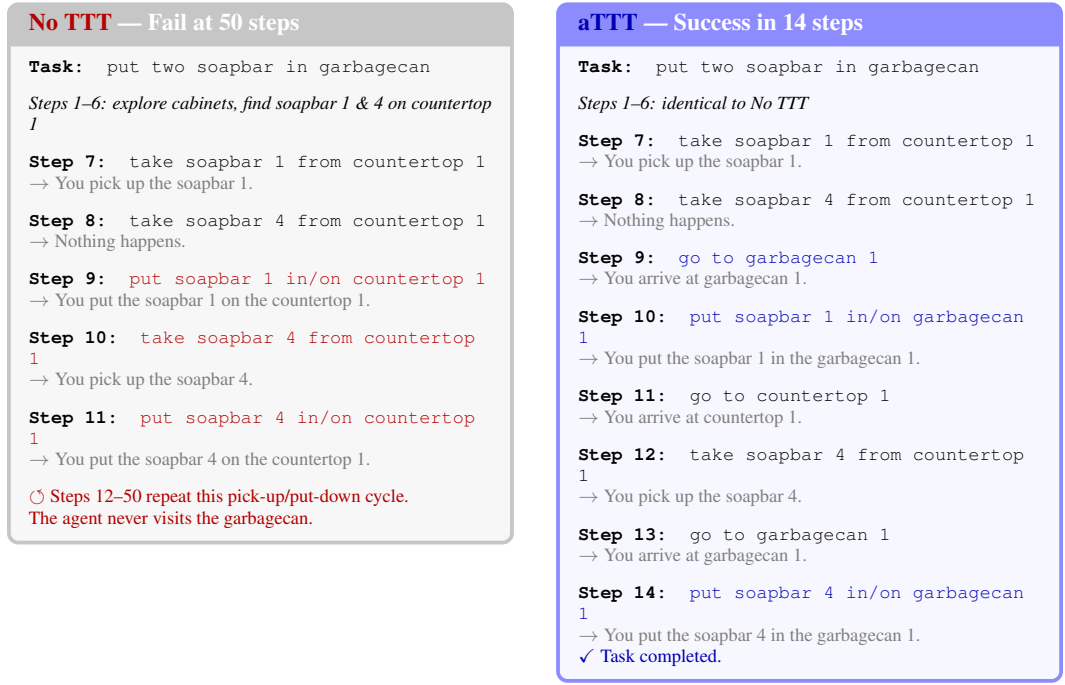

\centering
\begin{minipage}[t]{0.48\textwidth}\vspace{0pt}
\begin{tcolorbox}[colback=gray!6, colframe=gray!45, fonttitle=\bfseries\small, title={\textcolor{red!70!black}{No TTT} — Fail at 50 steps}, left=4pt, right=4pt, top=2pt, bottom=2pt]
\scriptsize\ttfamily
\textbf{Task:} put two soapbar in garbagecan\par\medskip
\textrm{\textit{Steps 1--6: explore cabinets, find soapbar 1 \& 4 on countertop 1}}\par\medskip
\textbf{Step 7:} take soapbar 1 from countertop 1\par
\textrm{\textcolor{gray}{$\rightarrow$ You pick up the soapbar 1.}}\par\medskip
\textbf{Step 8:} take soapbar 4 from countertop 1\par
\textrm{\textcolor{gray}{$\rightarrow$ Nothing happens.}}\par\medskip
\textbf{Step 9:} \textcolor{red!70!black}{put soapbar 1 in/on countertop 1}\par
\textrm{\textcolor{gray}{$\rightarrow$ You put the soapbar 1 on the countertop 1.}}\par\medskip
\textbf{Step 10:} \textcolor{red!70!black}{take soapbar 4 from countertop 1}\par
\textrm{\textcolor{gray}{$\rightarrow$ You pick up the soapbar 4.}}\par\medskip
\textbf{Step 11:} \textcolor{red!70!black}{put soapbar 4 in/on countertop 1}\par
\textrm{\textcolor{gray}{$\rightarrow$ You put the soapbar 4 on the countertop 1.}}\par\medskip
\textcolor{red!70!black}{\textrm{$\circlearrowleft$ Steps 12--50 repeat this pick-up/put-down cycle.}}\par
\textcolor{red!70!black}{\textrm{The agent never visits the garbagecan.}}
\end{tcolorbox}
\end{minipage}\hfill
\begin{minipage}[t]{0.48\textwidth}\vspace{0pt}
\begin{tcolorbox}[colback=blue!3, colframe=blue!45, fonttitle=\bfseries\small, title={\textcolor{blue!70!black}{aTTT} — Success in 14 steps}, left=4pt, right=4pt, top=2pt, bottom=2pt]
\scriptsize\ttfamily
\textbf{Task:} put two soapbar in garbagecan\par\medskip
\textrm{\textit{Steps 1--6: identical to No TTT}}\par\medskip
\textbf{Step 7:} take soapbar 1 from countertop 1\par
\textrm{\textcolor{gray}{$\rightarrow$ You pick up the soapbar 1.}}\par\medskip
\textbf{Step 8:} take soapbar 4 from countertop 1\par
\textrm{\textcolor{gray}{$\rightarrow$ Nothing happens.}}\par\medskip
\textbf{Step 9:} \textcolor{blue!70!black}{go to garbagecan 1}\par
\textrm{\textcolor{gray}{$\rightarrow$ You arrive at garbagecan 1.}}\par\medskip
\textbf{Step 10:} \textcolor{blue!70!black}{put soapbar 1 in/on garbagecan 1}\par
\textrm{\textcolor{gray}{$\rightarrow$ You put the soapbar 1 in the garbagecan 1.}}\par\medskip
\textbf{Step 11:} go to countertop 1\par
\textrm{\textcolor{gray}{$\rightarrow$ You arrive at countertop 1.}}\par\medskip
\textbf{Step 12:} take soapbar 4 from countertop 1\par
\textrm{\textcolor{gray}{$\rightarrow$ You pick up the soapbar 4.}}\par\medskip
\textbf{Step 13:} go to garbagecan 1\par
\textrm{\textcolor{gray}{$\rightarrow$ You arrive at garbagecan 1.}}\par\medskip
\textbf{Step 14:} \textcolor{blue!70!black}{put soapbar 4 in/on garbagecan 1}\par
\textrm{\textcolor{gray}{$\rightarrow$ You put the soapbar 4 in the garbagecan 1.}}\par
\textcolor{blue!70!black}{\textrm{\checkmark\ Task completed.}}
\end{tcolorbox}
\end{minipage}
\caption{Case study: \textit{put two soapbar in garbagecan} (game 100, Qwen3.5-9B). Both agents explore identically through step 8. At step 9, the no-TTT agent puts the soapbar back down and enters a 41-step loop. The aTTT agent carries it to the garbagecan and finishes in 14 steps.}
\label{tab:case}
\end{figure}   

\end{document}